\documentclass[conference]{IEEEtran}
\IEEEoverridecommandlockouts
\usepackage{cite}
\usepackage{amsmath,amssymb,amsfonts}
\usepackage{algorithmic}
\usepackage{graphicx}
\usepackage{textcomp}
\usepackage{subcaption}
\usepackage{booktabs}
\usepackage{stfloats}
\usepackage{float}
 \usepackage{url}
\usepackage{xcolor}
\def\BibTeX{{\rm B\kern-.05em{\sc i\kern-.025em b}\kern-.08em
    T\kern-.1667em\lower.7ex\hbox{E}\kern-.125emX}}
\begin{document}

\title{Interactive Instance Annotation with Siamese Networks}

\author{Xiang~Xu,~
        Ruotong~Li,~
        Mengjun~Yi,~
        Baile~XU,~
        Furao~Shen,~
        and~Jian~Zhao,}

\maketitle

\begin{abstract}
Annotating instance masks is time-consuming and labor-intensive. A promising solution is to predict contours using a deep learning model and then allow users to refine them. However, most existing methods focus on in-domain scenarios, limiting their effectiveness for cross-domain annotation tasks. In this paper, we propose SiamAnno, a framework inspired by the use of Siamese networks in object tracking. SiamAnno leverages one-shot learning to annotate previously unseen objects by taking a bounding box as input and predicting object boundaries, which can then be adjusted by annotators. Trained on one dataset and tested on another without fine-tuning, SiamAnno achieves state-of-the-art (SOTA) performance across multiple datasets, demonstrating its ability to handle domain and environment shifts in cross-domain tasks. We also provide more comprehensive results compared to previous work, establishing a strong baseline for future research. To our knowledge, SiamAnno is the first model to explore Siamese architecture for instance annotation.
\end{abstract}


\section{Introduction}\label{sec1}

In the era of deep learning, accurately annotated datasets are essential for tasks such as object detection, instance segmentation, and visual tracking.Many computer vision tasks have progressed from coarse bounding-box annotations to precise pixel-level annotations. For instance, research has shifted its focus from object detection to instance segmentation.

However, annotating ground truth instance masks is an extremely time-consuming and labor-intensive task. Previous research indicates that human annotators spend an average of 20-30 seconds per object\cite{polygonrnnpp}. 
Recent works aim to further reduce human effort by leveraging the deep learning (DL) models\cite{dec, iog, fca, polygonrnnpp, curvegcn, splitgcn}. Researchers have designed various segmentation networks along with different human-computer interaction mechanisms, and some have been extended to domains  such as cancer diagnoses\cite{gcnmr, liu2022transforming, du2022interactive}.

Instance annotation networks can be categorized into pixel-wise and contour-wise methods. While pixel-wise approaches excel in segmentation tasks, their binary masks are hardly modifiable. In contrast, contour-wise methods allow annotators to directly adjust predicted vertices, making them more user-friendly for interactive refinement. This paper introduces a contour-wise method designed to enhance annotator efficiency through intuitive interaction.
Although deep learning-based contour-wise models have shown promising results\cite{polygonrnn, polygonrnnpp, curvegcn, dacn, splitgcn}, they often focus on objects within the training set and neglect the challenges posed by domain and environmental shifts, which is an inevitable factor in annotation tasks. 

While some studies test models on different datasets, they often prioritize in-domain performance over generalization. They usually compare multiple metrics (e.g., mIoU, mAP, F score) for in-domain tasks but report only mIoU for cross-domain performance. To bridge this gap, this paper offers comprehensive evaluations across both scenarios, establishing a strong baseline for future research.

In addition, existing models do not take into account the generalization ability at the network design stage. All those methods simply utilize the one-pass convolutional networks as their backbones\cite{resnet}, whose outputs are further processed by a boundary prediction network. Such a design lacks the one-shot learning ability, resulting in inferior performance when annotating new datasets. 

In recent years, the Siamese architecture has become the standard approach for designing video object tracking (VOT) models\cite{uast}. Object annotation, which aims to separate foreground pixels from the background, shares similarities with VOT tasks, where Siamese networks track user-defined targets in each stand-alone frames. Inspired by the success of Siamese networks in tracking, we propose {\bf SiamAnno}, an architecture adapted for segmentation annotation.

{\bf SiamAnno} learns a segmentation network that converts bounding boxes into boundary contours. Since bounding boxes are easier and more cost-effective to obtain, they reduce annotators’ workload when creating polygon annotations. Image crops centered on the objects of interest are input into SiamAnno’s two branches to extract features. SiamAnno then fuses these features using pixel-wise correlation and estimates vertex positions along the object boundary via a contour prediction head. 
The contributions of this paper are summarized as follows:
\begin{itemize}
\item[$\bullet$] We explore Siamese architecture in the context of instance annotation, and obtain SOTA performances on multiple datasets.
\item[$\bullet$] We present experimental results using multiple metrics in the cross-domain tasks, to provide strong and thorough baselines for future studies. 
\end{itemize}

\section{Related Work}\label{sec2}
\subsection{Siamese Networks}
The general Siamese neural network\cite{bromley1993signature} consists of two branches that share identical architecture and the same weights. Two images, forming a pair, are passed through the same sub-networks, yielding two outputs that are then concatenated and passed on for further computation. It has been shown that the Siamese architecture fits any neural networks including CNNs, RNNs\cite{mueller2016siamese} and Restricted Boltzmann Machines\cite{nair2010rectified}. Siamese networks have driven advancements in fields with the ability of exploring the intrinsic similarity under the feature space. 

In recent years, the Siamese architecture has become the standard prototype to design new trackers in VOT tasks and has recorded several SOTA results\cite{uast}. In VOT, the tracker tracks \emph{any} target specified in the first frame by a human, which may not exist in the training set. Since the Siamese-based VOT models do not rely on movement continuity, these tracking models have to separate the probably previous-unseen object from the background individually in each static frame, which is similar to annotating new objects in stand-alone images. The need to annotate previously unseen objects is inevitable when working with new datasets. 


\subsection{Instance Annotation}

{\bf Pixel-wise Methods} frame segmentation tasks as per-pixel classification problems. Early methods \cite{grabcut} used graph cuts based on color and texture cues, while deep learning approaches have since outperformed traditional ones in accuracy. DEXTR \cite{dec} segments objects using four user-provided extreme points, while IOG \cite{iog} reduces the number of clicks by adding an interior click and two corner clicks. FCA-Net \cite{fca} highlights the importance of the first click as an anchor, further minimizing user interaction. Some methods, such as those involving coarse segmentation followed by detail refinement \cite{focalclick, lin2022multi} and boundary refinement modules \cite{dong2021lightweight, tang2021look} , produce more accurate boundaries.
However, generating binary masks, these methods require pixel-by-pixel adjustments for accurate edits \cite{wang2018deepigeos}, which is user-unfriendly and labor-intensive.

{\bf Contour-wise Methods} detect object contour curves that consist of verticves and edges. Level set segmentation\cite{levelset} frames object annotation as curve evolution, predicting object boundaries by continuously taking derivatives on the well-designed energy function of the curves.
DELSE\cite{delse} uses a CNN to predict the evolution parameters, making the level-set framework end-to-end trainable. However, users cannot directly drag the boundaries in these methods.


Intelligent Scissors\cite{intelsciss} allow users to trace the boundary by simply moving the mouse in proximity to the object's edge. 
Polygon-RNN\cite{polygonrnn} adopts a similar idea of sequentially predicting the vertices, but in a deep-learning way. Human corrections can be fed to the RNN to replace the model's prediction, helping the model to get back on the right track. Polygon-RNN++\cite{polygonrnnpp} improves both the network architecture and the training scheme, and increases the output resolution, but still suffers to slow reference time and low scalability of vertex numbers.
DACN\cite{dacn} further combines both the edge and segmentation features in a multi-task learning framework but shows a limited prediction of disconnected objects. The separating network in Split-GCN\cite{splitgcn} reconstructs the vertex topology to express the object’s shape containing the disconnected components. Our method follows their design that takes the image crops as the input and outputs the predicted vertices along the boundary. 

\section{Proposed Methods}\label{sec3}

\subsection{Overview}\label{sec:taskdescription}
\begin{figure*}[htbp]
  \centering
   \includegraphics[width=0.85\linewidth]{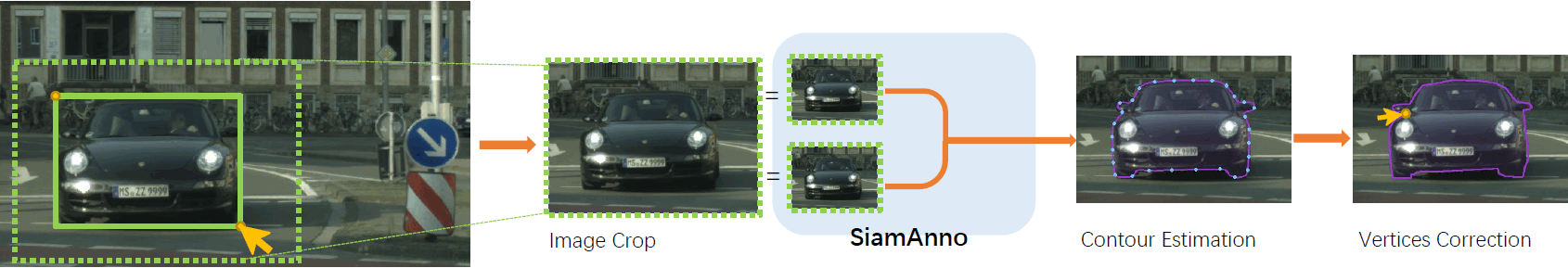}
   \caption{SiamAnno for instance annotation. Annotators wrap the instance by dragging a bounding box or input the bounding box predicted by some object detection model. SiamAnno takes the the bounded region with a certain slack as the input of its two branches, and outputs the predicted contour. Users can further correct the estimated boundary by pulling the vertices.}
   \label{fig:task_description}
\end{figure*}

The instance annotation process is illustrated in Fig. \ref{fig:task_description}. Our annotation model trains a segmentation network to convert the bounding box of an object into contours represented by vertices. The bounding box input can come from users’ real-time interactions or existing dataset annotations. The output vertices should be interactive, allowing annotators to modify them manually. 

The model is trained on a dataset $D_{train}$ and evaluated on another dataset $D_{test}$. When $D_{train}$ and $D_{test}$ share the same distribution (e.g., train and validation splits of the same dataset), it is regarded as an \emph{in-domain} annotation task. For the \emph{cross-domain} scenario, $D_{test}$ contains different objects and backgrounds. 

SiamAnno is a segmentation network that converts image crops into boundary contours efficiently, supports for interactive adjustment of results and shows great potential especially in cross-domain annotation tasks. Specifically, SiamAnno can handle zero-shot annotation without retraining or fine-tuning. 


\subsection{Network Architecture}\label{sec:method}

The Siamese network consists of two input branches and a feature fusion module that generates a correlation map. The contour prediction head then estimates vertex positions. To improve regression accuracy, a U-Net-style feature fusion mechanism is added before the prediction head to leverage multi-level features.

\begin{figure*}
  \centering
   \includegraphics[width=0.85\linewidth]{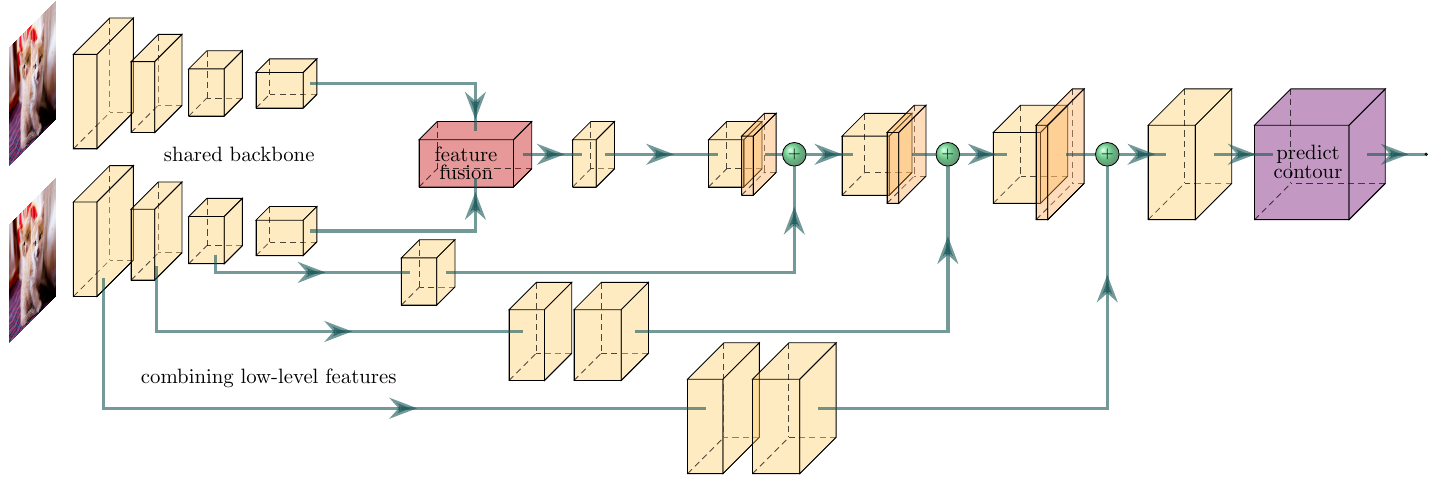}
   \caption{The network architecture of SiamAnno. Features from the search branch and the target branch are correlated to produce a fused feature map. We expand the feature map by combining low-level features and send it to the contour prediction head.}
   \label{fig:ardnet}
\end{figure*}

\noindent {\bf Siamese-based Feature Extraction.} As illustrated in Fig. \ref{fig:ardnet}, our model consists of two input branches: the \emph{target branch} and the \emph{search branch}. Both branches process regions containing the instance of interest, with the entire image cropped into a concentric search region, scaled by a factor of $s$ times the ground-truth bounding box size (referred to as the \emph{search scale}). Two crops are passed through the respective branches, utilizing a parameter-shared backbone to extract features. Then the feature maps from the two branches are input to the correlation module at different scales. The target branch only uses the central $\frac{1}{s^2}$ area of the feature map, while the search branch retains its full-size feature map. Such a strategy enables the target branch focus on the target, while the search branch captures additional background clues. 

\noindent {\bf Pixel-wise Correlation.} 
In most VOT methods, the search region is four times the size of the target ($s=4$). While in our case, the scale factor $s\in(1,2)$. Comparing to widely used naive correlation or depth-wise correlation, pixel-wise correlation is better at maintaining spatial information when the feature maps of the two branches have small differences in size.

Pixel-wise correlation takes each pixel as a kernel. For the target branch feature $T\in\mathbb{R}^{c\times H_{t}\times W_{t}}$ and the search branch feature $S\in\mathbb{R}^{c\times H_{s}\times W_{s}}$, pixel-wise correlation decomposes $T$ into kernels $k_{i}\in\mathbb{R}^{c\times1\times1}$ and computes correlation separately on each kernel to obtain the correlation map $C\in\mathbb{R}^{H_{t} \times W_{t} \times H_{s} \times W_{s}}$:
\begin{equation}\label{eq:correlation}
    C=\{C_{i}|C_{i}=k_{i}*S\}_{i\in\{1,...,H_{t}W_{t}\}},
\end{equation}
where $*$ is the naive correlation.

We implement a U-net style feature fusion mechanism to enlarge the correlation map and facilitate the subsequent contour prediction. Specifically, we use feature maps derived from convolutional layers of the search-branch backbone in inverse order. We define the usage of one feature map as a step. In step $j$, the fused feature map $M_{j-1}$ from the previous step is first interpolated into twice its resolution and added with the corresponding backbone feature map $C_{j}$. The derived feature map is then passed through a convolution layer and a ReLU function and finally sent into the next step.
\begin{equation}\label{eq:upsample}
    M_{j} = f_{j}(\mathrm{Interpolate}(M_{j-1})+C_{j}).
\end{equation}

The aforementioned correlation map is passed through a convolutional layer to get the initial $M_0$. 
Features from later layers contain high-level semantic information, while the earlier layers produce features with low-level information such as color and shape, which are essential for precise boundary detection. Feature maps expanded from such a coarse-to-fine fusion mechanism retain hierarchical information, enabling the contour prediction head to estimate vertices at high resolution.

\subsection{Contour Prediction Head}
The prediction head estimates the offset of each vertex, to shrink the initial bounding box to the precise object boundary. 
We sample $K$ points, regard as the initial boundary, along the bounding box. Circular convolution is applied to each vertex, where the input consists of per-vertex features extracted from the entire feature map, and the output is the corresponding offset. 
Eq. \eqref{eq:circular} defines the circular convolution on vertex $p$:
\begin{equation}\label{eq:circular}
    (f*k')_{p} = \sum^{R}_{r=-R}f_{p+r}k'_{r}
\end{equation}
where $f$ is the feature map, $k'$ is the learnable kernel function, $*$ is the standard 1D convolution and $R$ is the size of the convolutional kernel. Such circular convolution can also be defined in a dilated way by defining the dilation rate, as the well-known dilated convolution. Cascading multiple dilated circular convolution with different dilation rate allows our method aggregates the features of both the nearest neighbor as well as the vertices in a certain distance, utilizing multi-scale boundary information.

When user adjust the result by dragging vertices, the updated boundary will be iteratively refined by feeding back into the pipeline. Subsequent iterations focus on inaccurately predicted vertices, using the attentive deformation mechanism\cite{dance} which outputs per-pixel modulation coefficients to adaptively reweigh the newly predicted offsets and the original estimated.



Four 1$\times$1 convolution layers are applied to the output from the cascaded circular convolutions with a \emph{tanh} function, producing the final vertex offset estimation. Note that the feature map used in Eq. \eqref{eq:circular} consists of both the learning-based features and the vertex coordinates. We normalize each coordinate by subtracting the minimum value over all boundary vertices, then dividing the horizontal/vertical coordinate by object's width/height. Such normalization converts the original coordinate into a relative one, making it scale and translation invariant, and helping stabilize the training process. 


\subsection{Loss Function}

We employ smooth $L_1$ loss to supervise the deformation at each vertex, as Eq. \eqref{eq:snakeloss} shows,
\begin{equation}\label{eq:snakeloss}
    L_{vertex} = \frac{1}{N}\sum_{n=1}^{N}smooth\_L_1\left(\frac{\tilde{x}_{n}}{W}-\frac{x_{n}}{W} \right)
\end{equation}
where the losses are averaged by the number of vertices $N$ on one boundary. $x$ is the ground truth vertex location and $\tilde{x}$ denotes the estimated vertex coordinate. Large objects produces larger estimation gap. We use the side length $W$ of each bounding box to weight the loss and stabilize the training.

The computation of $L_{vertex}$ needs one-by-one correspondence between the estimated contour points and the target points. We apply the segment-wise matching scheme introduced in \cite{dance}, where the intersection points of the ground-truth object boundary and the initial contour (the bounding box) split the entire ground-truth contour into multiple segments. The assignment of ground-truth vertices is performed locally within each segment, so as to relieve the correspondence interlacing phenomenon in the previous methods and smooth the learning process. We refer the interested readers to \cite{dance} for details. We use Dice loss\cite{diceloss} in training the attentive deformation mechanism, combing $L_{Dice}$ and the regression loss $L_{vertex}$ into the overall loss $L=L_{Dice}+\alpha L_{vertex}$. In our implementation, we set $\alpha=10$ by default.


\section{Experiments}
\label{sec:experiments}
In this section, we conduct comprehensive experiments to verify SiamAnno's effectiveness on both in-domain and cross-domain annotations tasks. 
Previous works usually focus on the former scenario, and make comprehensive comparisons using multiple metrics. However, when it comes to the cross-domain scenario, only the mIoU metric is employed. We hold the opinion that annotating new objects is also of great importance, and its evaluations should be done more sufficiently. In cross-domain tasks, we will not only report the mIoU measures as previous works did, but also the mAP and the boundary F scores. We hope to provide a baseline for future studies.

\subsection{Evaluation Metric}
The intersection over union (IoU) metric is first computed on a per-instance basis, then averaged in each category. As with the previous work\cite{polygonrnn, polygonrnnpp, curvegcn, splitgcn}, the reported mIoU is the average over these per-category IoU scores, not over the original IoUs of each instance. 

Average Precision (AP) is a widely used metric in segmentation. Different from the computation of mIoU, the AP is computed across all instances in previous works\cite{curvegcn}, with no consideration of the categories. We follow this methodology. We compute mAP by increasing the IoU overlap threshold from 0.5 to 0.95 with a step of 0.05, and report the average of them, which is usually denoted as mAP@(0.5:0.95) in instance segmentation literatures. 

Both IoU and AP are computed by comparing the difference in area between the prediction and the ground truth, without considering contour accuracy. The boundary F score measures the precision and recall by counting the hits, misses, and false positives based on a correspondence of machine and human boundary pixels matched by morphology operators. Small localization errors are permitted by controlling the tolerable pixel numbers, and we report results at thresholds of 1 and 2 pixels as in \cite{curvegcn, splitgcn}, denotes as $\mathrm{F}_{1px}$ and $\mathrm{F}_{2px}$ respectively.

\subsection{Comparisons with State-of-the-Arts}
\subsubsection{In-Domain Annotation}\label{sec:indomain}

\begin{table*}[ht]
  \centering
  \begin{tabular}{lcccccccc|c}
    \toprule
    Model           & Bicycle & Bus & Person & Train & Truck & Motorcycle & Car & Rider & \bf{mIoU} \\
    \midrule
    Polygon-RNN     & 52.13 & 69.53 & 63.94 & 53.74 & 68.03 & 52.07 & 71.17 & 60.58 & 61.40 \\
    Polygon-RNN++   & 63.06 & 81.38 & 72.41 & 64.28 & 78.90 & 62.01 & 79.08 & 69.95 & 71.38 \\
    DACN            & 64.58 & 82.60 & 72.93 & 61.25 & 80.51 & 63.85 & 80.31 & 71.29 & 72.17 \\
    Polygon-GCN     & 64.55 & 85.01 & 72.94 & 60.99 & 79.78 & 63.87 & 81.09 & 71.00 & 72.66 \\
    PSP-DeepLab     & 67.18 & 83.81 & 72.62 & 68.76 & 80.48 & \bf{65.94} & 80.45 & 70.00 & 73.66 \\
    Spline-GCN      & \bf{67.36} & \bf{85.43} & \bf{73.72} & 64.40 & 80.22 & 64.86 & \bf{81.88} & \bf{71.73} & 73.70 \\
    DELSE           & 67.15 & 83.38 & 73.07 & 69.10 & \bf{80.74} & 65.29 & 81.08 & 70.86 & \bf{73.84} \\
    \midrule
    SiamAnno\hspace{2pt} (Ours)  & 63.89 & 80.61 & 72.12 & \bf{70.25} & 80.11 & 64.02 & 79.40 & 68.19 & 72.33 \\
    SiamAnno\dag(Ours) & \bf{\emph{69.11}} & \emph{85.26} & \bf{\emph{75.37}} & \bf{\emph{77.79}} & \bf{\emph{82.54}} & \bf{\emph{69.80}} & \bf{\emph{82.66}} & \emph{71.02} & \bf{\emph{76.69}} \\
    \bottomrule
  \end{tabular}
  \caption{In-domain performances (IoU in \% in val test) on all the Cityscapes categories. 
  }
  \label{tab:cityscapes_miou}
  
\end{table*}

\begin{table}[t]
  \centering
  \begin{tabular}{lccc}
    \toprule
    Model           & mAP & $\mathrm{F}_{1px}$ & $\mathrm{F}_{2px}$ \\
    \midrule
    DACN            & -     & 45.27 & 59.89 \\
    Polygon-RNN++   & 25.5  & 46.57 & 62.26 \\
    PSP-Deeplab     & -     & 47.10 & 62.82 \\
    Spline-GCN      & -     & 47.72 & 63.64 \\
    DELSE           & -     & 48.59 & 64.45 \\
    Split-GCN       & 29.6  & \bf{52.50} & \bf{67.50} \\
    \midrule
    SiamAnno\hspace{2pt} (Ours) & \bf{39.6} & 46.62 & 60.20 \\
    SiamAnno\dag(Ours) & \bf{\emph{48.5}} & \emph{52.43} & \emph{66.68} \\
    \bottomrule
  \end{tabular}
  \caption{In-domain performance in terms of mAP and F score on Cityscapes.}
  \label{tab:cityscapes_map}
\end{table}

\begin{table}[t]
  \centering
  \begin{tabular}{lccc}
    \toprule
    Model           & KITTI & ADE20k & Rooftop \\
    \midrule
    Polygon-RNN     & 74.22 & -     & -     \\
    Polygon-RNN++   & 83.14 & 71.82 & 65.67 \\
    PSP-Deeplab     & 83.35 & 72.70 & 57.91 \\
    Polygon-GCN     & 83.66 & 72.31 & 66.78 \\
    Spline-GCN      & 84.09 & 72.94 & 68.33 \\
    DACN            & -     & 73.21 & 66.92 \\
    SiamAnno (Ours)  & \bf{86.41} & \bf{74.90} & \bf{78.04} \\
    \bottomrule
  \end{tabular}
  \caption{Cross-domain performances (mIoU in \% in val test) on KITTI, ADE20k and Rooftop.}
  \label{tab:crossdomain}
\end{table}

\begin{table}[t]
  \centering
  \begin{tabular}{lcccc}
    \toprule
    Dataset & mIoU & mAP & $\mathrm{F}_{1px}$ & $\mathrm{F}_{2px}$ \\
    \midrule
    \multicolumn{5}{l}{- Train on COCO} \\
    COCO  & 79.85 & 56.3 & 59.17 & 71.37 \\
    \midrule
    \multicolumn{5}{l}{- Train on Cityscapes}\\
    Cityscapes & 72.33 & 39.6 & 46.62 & 60.20 \\
    Cityscapes$\dag$ & 76.69 & 48.9 & 52.88 & 67.90 \\
    KITTI    & 86.41 & 69.3 & 67.56 & 81.37 \\
    ADE20k   & 74.90 & 46.6 & 58.87 & 73.27 \\
    Rooftop  & 78.04 & 49.9 & 27.76 & 40.01 \\
    \bottomrule
  \end{tabular}
  \caption{SiamAnno's performances on different train/test combination. }
  \label{tab:crossdomain_more}
\end{table}

The Cityscapes\cite{cityscapes} dataset consists of street scenery images taken from 27 European cities. It has been split into 2975 training, 500 validation, and 1525 testing images. Since we do not have ground truth annotations on the test set, we follow the implementation in previous works\cite{polygonrnn, polygonrnnpp, curvegcn, splitgcn}, train our model on the train set, and report the results on the validation set. 

The dataset contains annotations for eight object categories, with significant size variance. Table \ref{tab:cityscapes_miou} reports the average IoU for each category, followed by their average as final mIoU scores, in line with previous works. Additionally, Table \ref{tab:cityscapes_map} presents the mAP and F scores. SiamAnno achieves an mAP of 39.6\%, marking a significant improvement, while its mIoU and F scores are competitive with existing methods. For instance, it performs best in annotating trains. The dataset is known for its fragmented instances, and our method, which deforms the estimated boundary to shrink the bounding box around the object, struggles with these due to the absence of a splitting mechanism for connected vertices. 
 Similar limitations are observed in other methods that model contours as cycle graphs \cite{polygonrnnpp, curvegcn}. As a result, SiamAnno does not outperform previous SOTA methods in in-domain tasks. 
 
 In a real labeling scenario, annotators can label separate components individually instead of enclosing a fragmented instance within a single bounding box.To simulate this, we also report SiamAnno’s performance in the per-component mode (marked with a$\dag$ in Table \ref{tab:cityscapes_miou} and \ref{tab:cityscapes_map})
 
 Changing from per-instance mode to per-components mode improves the performance in all metrics by a large margin,  especially the mAP, which has boosted from 39.6\% to 48.5\%, which leads to substantial improvements across all metrics, particularly mAP, which rises from 39.6\% to 48.5\%.


\subsubsection{Cross-Domain Annotation}
\label{sec:crossdomain}

KITTI\cite{kitti} is asmaller urban scene dataset compared to the Cityscapes. Images are captured in different cities, allowing us to test our model's ability to handle the environment shifts. Following \cite{polygonrnn, polygonrnnpp, curvegcn}, we use a derivative version of the dataset\cite{segkitti} and focus on the annotations of cars only.

ADE20k\cite{ade20k} is a general scene image segmentation dataset with a wide range of scenes and object categories with dense and detailed annotations. 
For a fair comparison with \cite{polygonrnnpp, curvegcn, dacn}, we select the following subset of categories: \emph{television receiver, bus, car, oven, person and bicycle}, and evaluate our method on the validation set.

Rooftop\cite{rooftop} contains 65 aerial images of rural scenes, differing from Cityscapes in the object category and the viewpoint. A majority of the building rooftops exhibit complex polygonal geometry, making the dataset a good test for model's capability in handling domain shift. Performance for this dataset is reported on the test set.

Table \ref{tab:crossdomain} shows the comparison with \cite{polygonrnn, polygonrnnpp, deeplab, curvegcn, dacn} on the above datasets in terms of the mIoU metric. Our approach consistently and significantly surpasses all other methods, which proves SiamAnno's great potential in handling the shift in environment and objects in the cross-domain annotation. 
To show more comprehensive results and provide a baseline for future studies, we also report the performances using the mAP metric and the F score in Table \ref{tab:crossdomain_more}.
The qualitative results obtained under the in-domain and the cross-domain annotation task intuitively demostrated that even facing the environment shift and domain shift, our SiamAnno model still produces high-quality contour prediction results. 
\begin{figure}[H]
  \centering
  \begin{subfigure}{0.32\linewidth}
    \includegraphics[width=\linewidth]{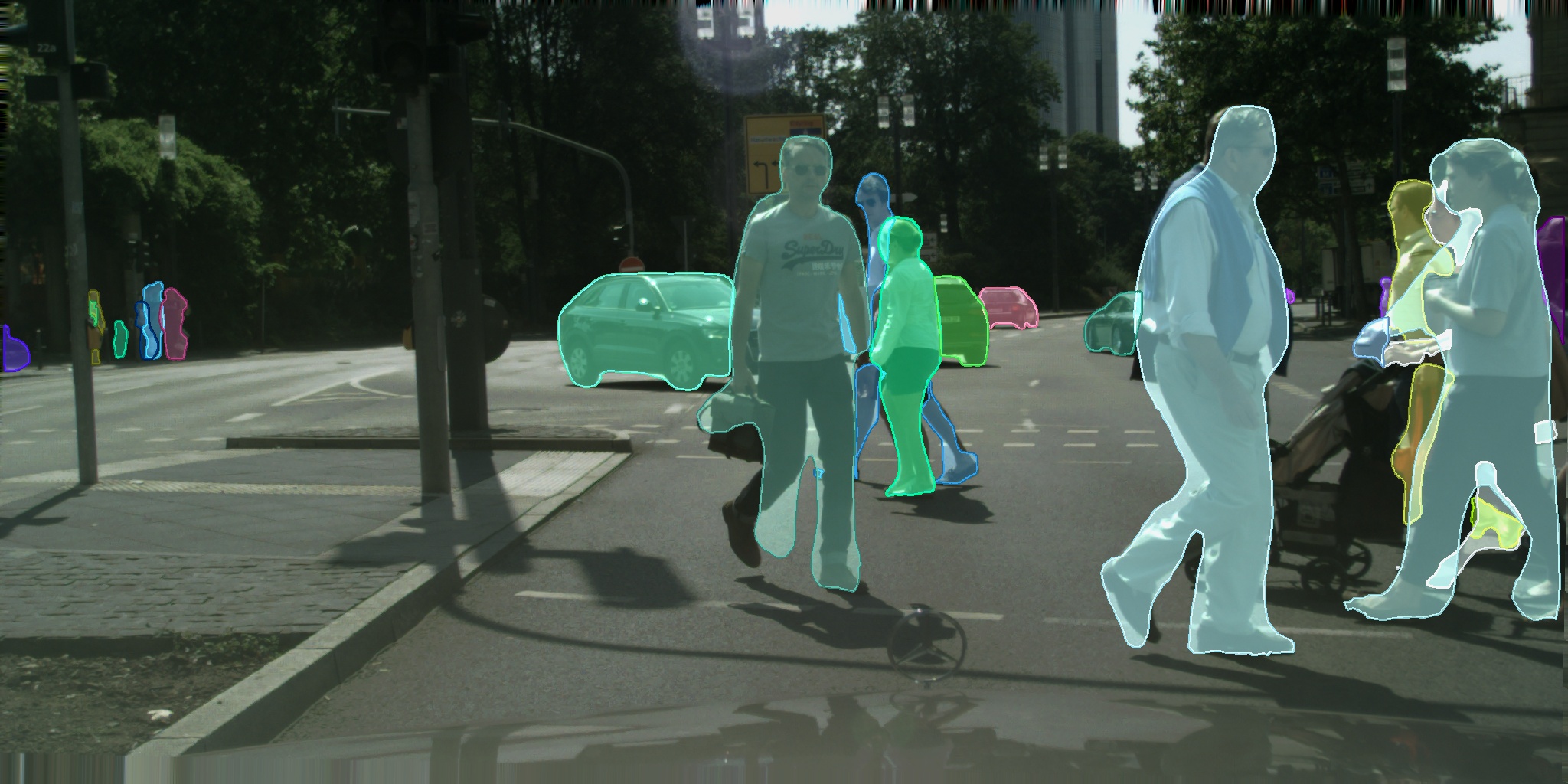}
  \end{subfigure}
  \begin{subfigure}{0.32\linewidth}
    \includegraphics[width=\linewidth]{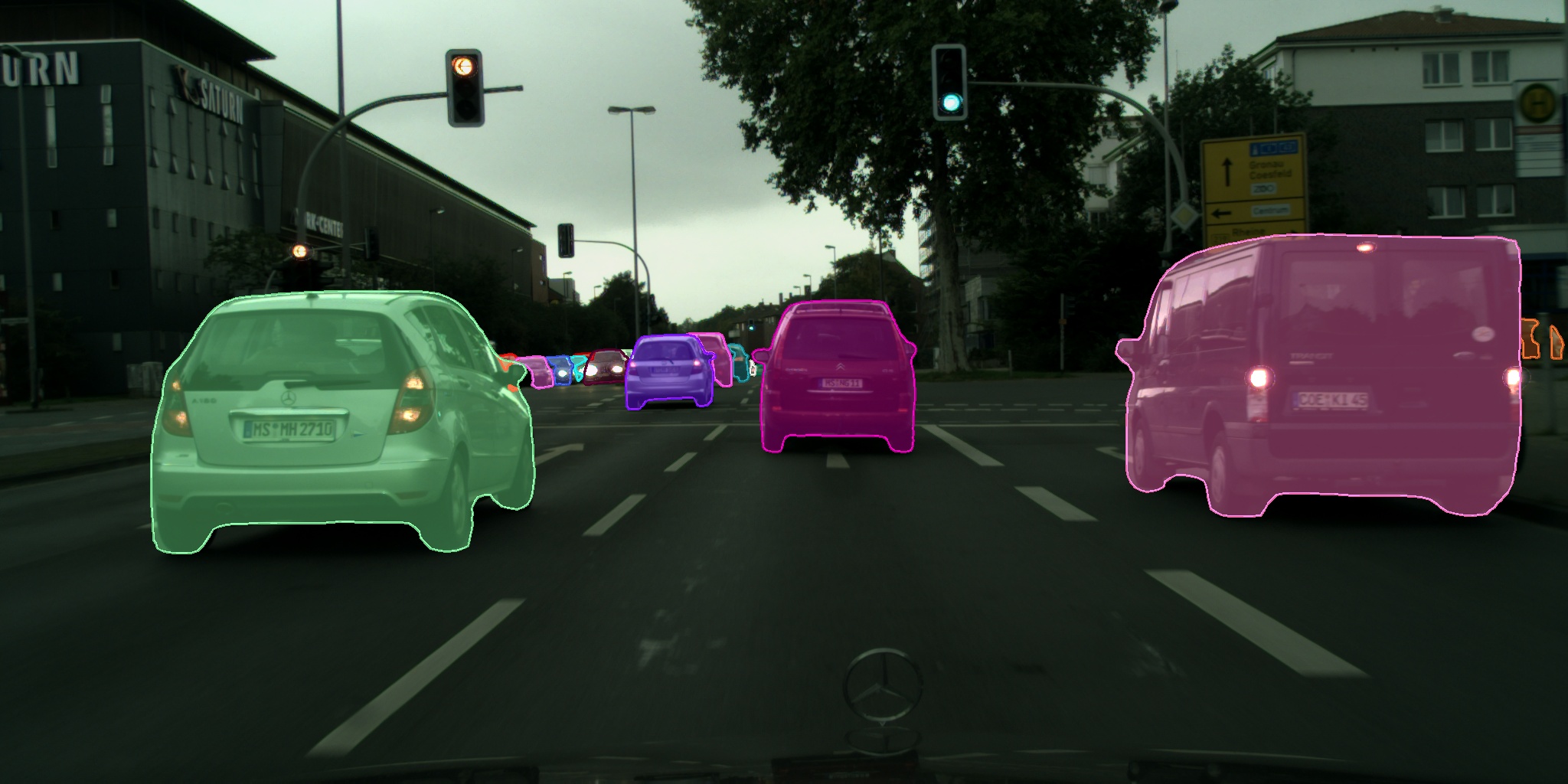}
  \end{subfigure}
  \begin{subfigure}{0.32\linewidth}
    \includegraphics[width=\linewidth]{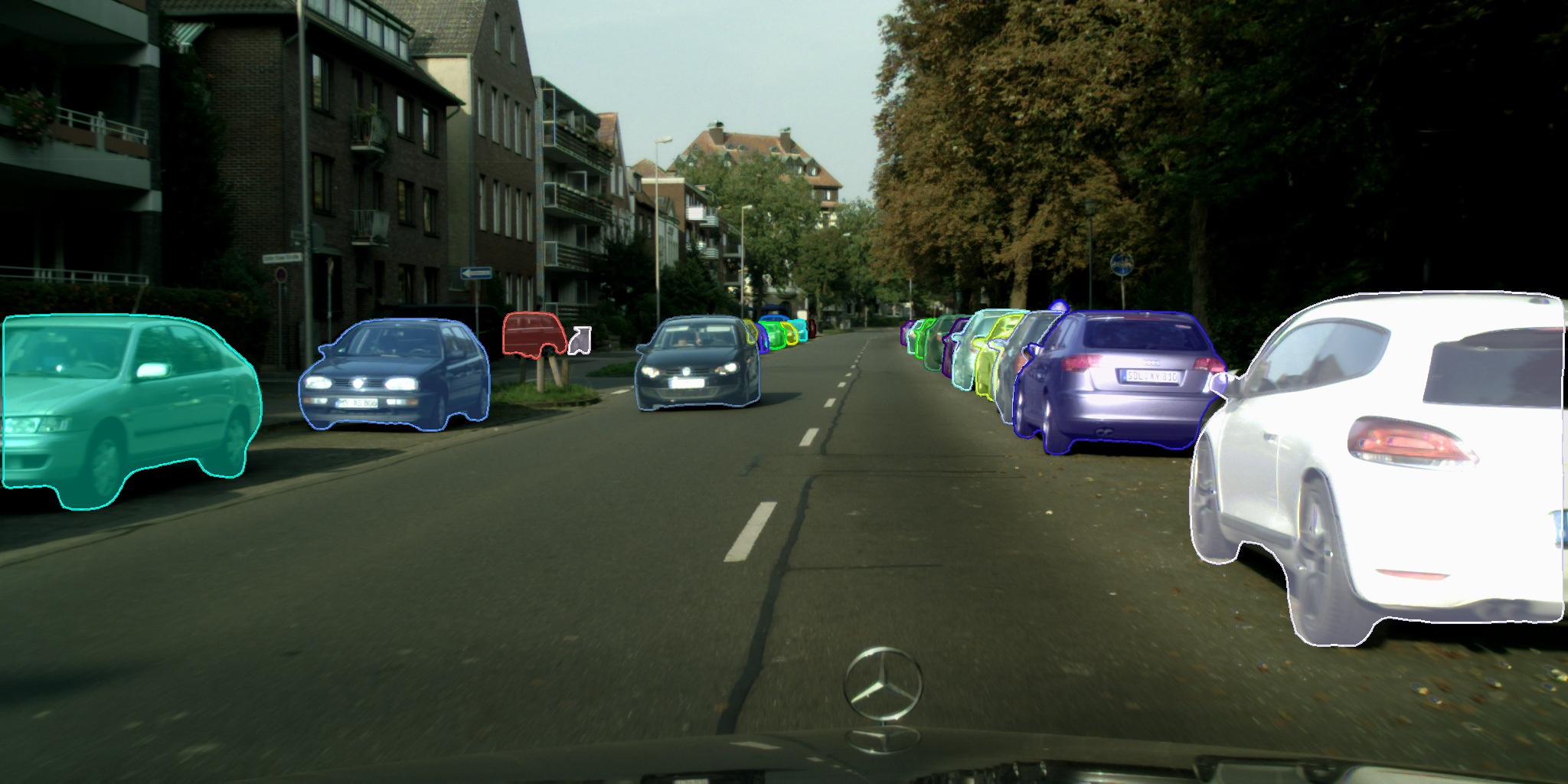}
  \end{subfigure}
   \caption{
   In-domain annotation results on Cityscapes. 
   Compared to the \emph{car}, inaccurate contours usually happen to the \emph{person} who may have irregular shape or movement (the leftmost image).}
   \label{fig:cityscapes_entireimg}
\end{figure}

\begin{figure}[H]
  \centering
  \begin{subfigure}{0.5\linewidth}
    \begin{minipage}[b]{\linewidth}
        \includegraphics[width=\linewidth]{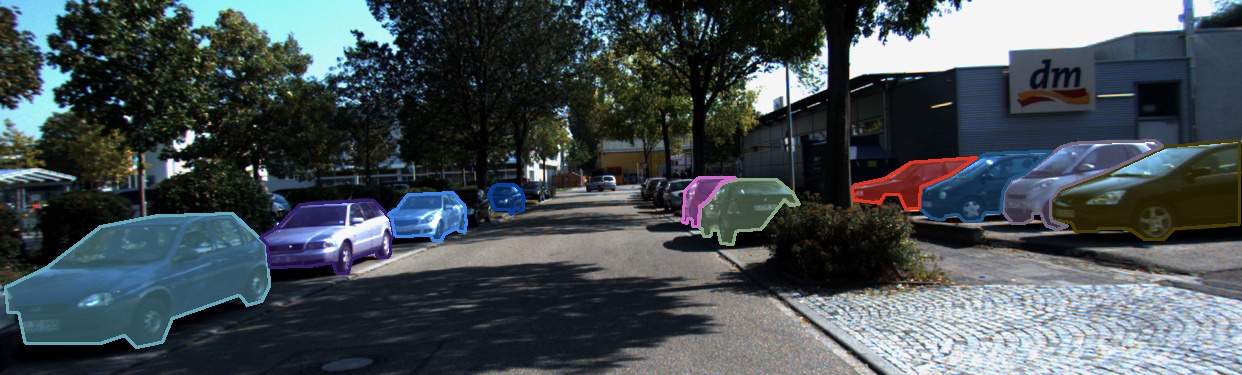}\vspace{2pt}
        \includegraphics[width=\linewidth]{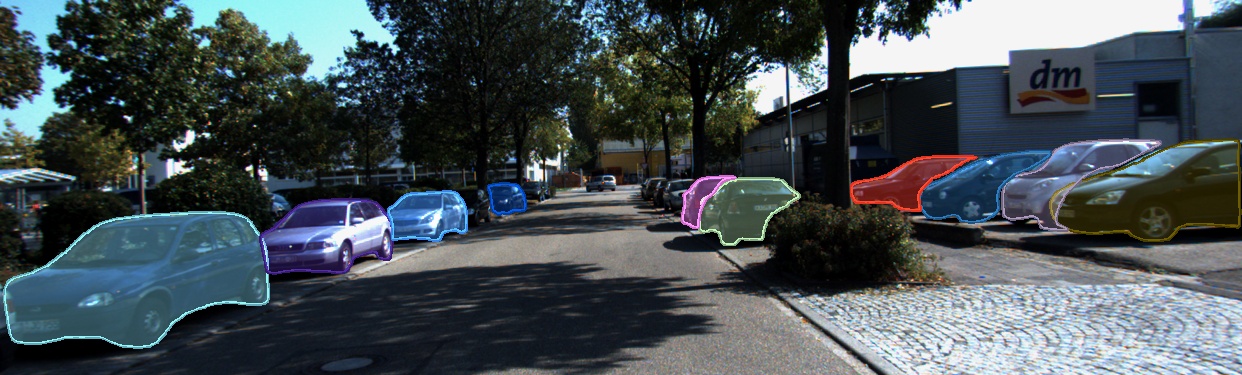}
    \end{minipage}
  \end{subfigure}
  \begin{subfigure}{0.22\linewidth}
    \begin{minipage}[b]{\linewidth}
        \includegraphics[width=\linewidth]{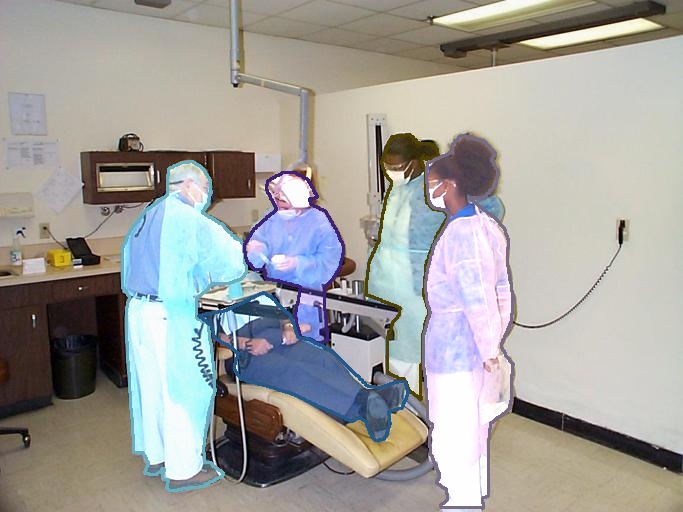}\vspace{2pt}
        \includegraphics[width=\linewidth]{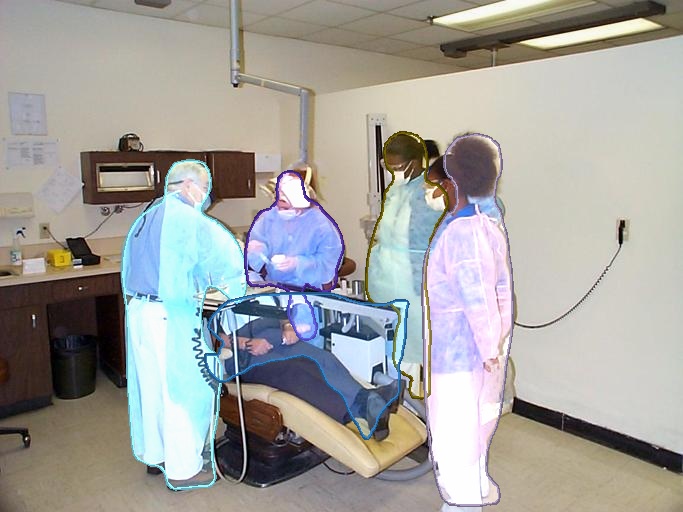}
    \end{minipage}
  \end{subfigure}
  \begin{subfigure}{0.22\linewidth}
    \begin{minipage}[b]{\linewidth}
        \includegraphics[width=\linewidth]{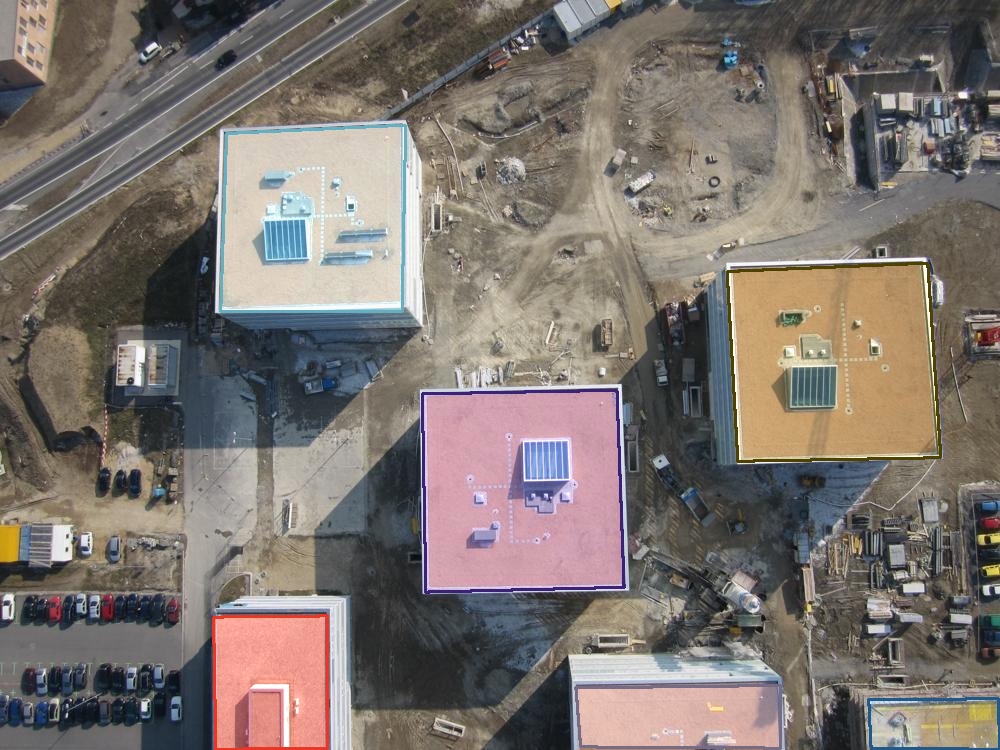}\vspace{2pt}
        \includegraphics[width=\linewidth]{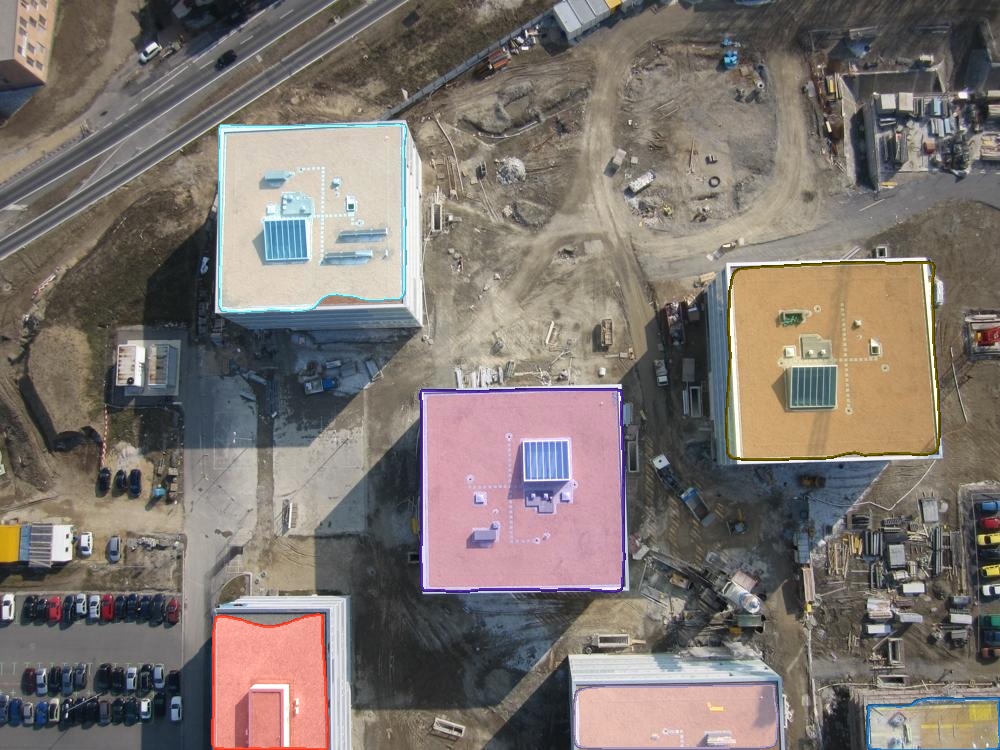}
    \end{minipage}
  \end{subfigure}
   \caption{Cross-domain annotation results on KITTI (the first column), ADE20k (the second column) and Rooftop (the third column). Note that the model here are only trained on Cityscapes without training or finetuning on these datasets. 
   }
   \label{fig:cross_domain}
\end{figure}

\begin{figure}[H]
  \centering
  \begin{subfigure}{0.49\linewidth}
    \includegraphics[width=\linewidth]{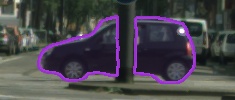}
  \end{subfigure}
  \begin{subfigure}{0.49\linewidth}
    \includegraphics[width=\linewidth]{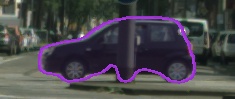}
  \end{subfigure}
  \caption{Comparison between the per-component mode (left) and the per-instance mode (right). Due to its nature of boundary shrinking, SiamAnno is not good at regressing object boundaries with breaks. }
   \label{fig:mode_comparison}
\end{figure}



\section{Conclusion}
\label{sec:conclusion}

We propose a Siamese-designed segmentation network, SiamAnno, for instance annotation. SiamAnno exploits the one-shot learning capability of the Siamese architecture, and the adapted snake-based contour prediction head accurately estimates vertex locations along the boundary. Experiments on ADE20k, KITTI, and Rooftop show that SiamAnno outperforms all previous methods and shows great potential in tackling environment shift and annotating previous-unseen objects. We also develop an annotation tool based on SiamAnno to facilitate segmentation annotation and the interested readers are referred to supplementary material for detail. Future work may include designing an annotator-in-the-loop correction mechanism so that the model can take advantage of user's modifications and re-predict the object boundary, further reducing alleviates human workload. 

\bibliographystyle{IEEEtran}
\bibliography{egbib}


\end{document}